\pdfoutput=1

\documentclass{article} %

\usepackage{acl}

\usepackage{times}
\usepackage{latexsym}

\usepackage[T1]{fontenc}

\usepackage[utf8]{inputenc}

\usepackage{microtype}

\usepackage{inconsolata}

\PassOptionsToPackage{numbers, compress}{natbib}

\usepackage[utf8]{inputenc} %
\usepackage[T1]{fontenc}    %
\usepackage{hyperref}       %
\usepackage{url}            %
\usepackage{booktabs}       %
\usepackage{amsfonts}       %
\usepackage{nicefrac}       %
\usepackage{microtype}      %
\usepackage{xcolor}         %

\usepackage{xspace}
\usepackage{graphicx}
\usepackage[frozencache]{minted}
\usepackage{soul}
\usepackage{tabularray}
\usepackage{enumitem}
\setlist{nosep}
\usepackage{tikz}
\newcommand{\twiceinclude}[2][]{%
  \begin{tikzpicture}%
    \node at (0,0) {\includegraphics[#1]{#2.png}};%
    \node at (0,0) {\includegraphics[#1]{#2.pdf}};%
  \end{tikzpicture}%
}%

\UseTblrLibrary{booktabs}
\newmintinline[code]{html}{}
\definecolor{ForestGreen}{RGB}{34,139,34}

\newcommand{\datasetname}{\textsc{ShopCART}}

\newcommand{\highlightRelevantValue}[1]{%
  \ifdim #1pt > 0pt
    {\color{ForestGreen}{#1\%}}%
  \else
    {\color{red}{#1\%}}%
  \fi
}

\title{Can LLM Agents Simulate Multi-Turn Human Behavior? Evidence from Real Online Customer Behavior Data}

\author{
  \textbf{Yuxuan Lu\textsuperscript{1}},
  \textbf{Jing Huang\textsuperscript{2}},
  \textbf{Yan Han\textsuperscript{2}},
  \textbf{Bingsheng Yao\textsuperscript{1}},
  \textbf{Sisong Bei\textsuperscript{2}},
  \textbf{Jiri Gesi\textsuperscript{2}},
  \\
  \textbf{Yaochen Xie\textsuperscript{2}},
  \textbf{Yisi Sang\textsuperscript{2}},
  \textbf{Zheshen (Jessie) Wang\textsuperscript{2}},
  \textbf{Qi He\textsuperscript{2}},
  \textbf{Dakuo Wang\textsuperscript{1}}
\\
\\
  \textsuperscript{1}Northeastern University,
  \textsuperscript{2}Amazon.com, Inc.
\\
  \small{
    \textbf{Correspondence:} \href{mailto:lu.yuxuan@northeastern.edu}{lu.yuxuan@northeastern.edu},
    \href{mailto:d.wang@northeastern.edu}{d.wang@northeastern.edu}
  }
}

\begin{document}
\maketitle
\begin{abstract}

Recent research shows that LLM Agents can generate ``believable'' human behaviors via prompt-only methods, and such agents have been increasingly adopted in downstream applications.
However, existing evaluation of these agents only focuses on qualitative believability (whether human raters think they are accurate), leaving open questions of whether LLM agents can accurately generate step-by-step actions mimicking a particular human's behavior in a multi-turn interaction task.
In this work, we take shopping as a case study and present the first large-scale quantitative evaluation of state-of-the-art LLMs' ability to accurately simulate human behavior.
Using real-world data from 31,865 online shopping sessions containing 230,965 user actions, our evaluation reveals that prompt-based LLMs (DeepSeek-R1, Llama, Claude) achieve only 11.86\% accuracy in generating human actions, highlighting \textbf{a substantial gap in actual behavioral accuracy}.
Through experiments, we also showcase that strategies as simple as fine-tuning LLMs on real human click-through data augmented with synthesized reasoning traces can greatly enhance models' performance. The fine-tuned Qwen2.5-7B achieves 17.26\% action generation accuracy and 33.86\% F1 score on final purchase prediction, representing substantial improvements of 5.4\% and 13.85\% over prompt-only baselines.
This work establishes the first rigorous benchmark for human behavior simulation and provides actionable insights for developing more accurate LLM agents for future downstream applications.

\end{abstract}

\begin{figure}[t]
    \centering
    \twiceinclude[width=\linewidth]{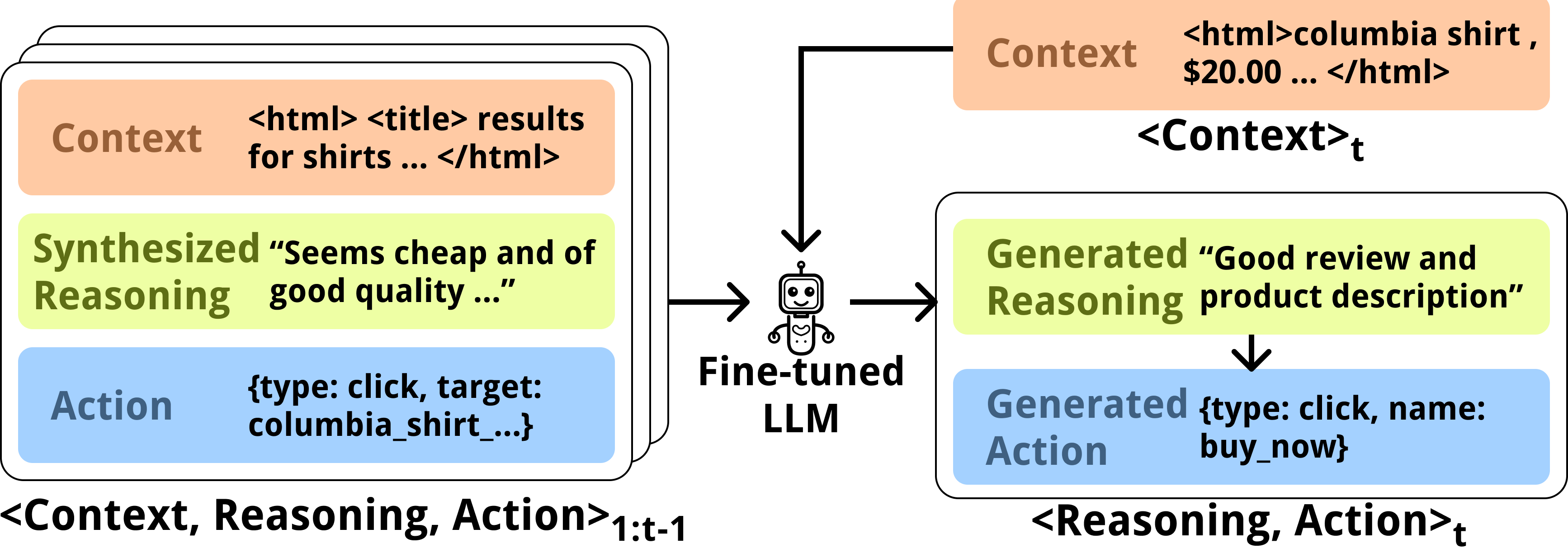}
\caption{Overview of the web action generation task. The model takes the currently observed \textbf{<context>$_{t}$} and a sequence of previous \textbf{<context, reasoning, action>$_{1:t-1}$} as input, and generates the next \textbf{<reasoning, action>$_{t}$} as output. Because the real-world human behavior dataset does not have groundtruth reasoning, we generate \textbf{synthesized reasoning trace} to complement the <context, action> pair. }
    \label{fig:model-arch}
    \vspace{-\baselineskip}
\end{figure}

\section{Introduction}

Recent advances in Large Language Models (LLMs) have enabled the simulation of human behavior across a range of applications, including web automation~\cite{gurRealWorldWebAgentPlanning2023, zhouWebArenaRealisticWeb2024}, social interaction behaviors~\cite{parkGenerativeAgentsInteractive2023,wu2025collabllm}, interpersonal trust behaviors~\cite{xieCanLargeLanguage2024}, and user interface interactions~\cite{taebAXNavReplayingAccessibility2024}. 
These developments have sparked a growing interest in developing \textbf{LLM Agents}.
Specifically, in the online shopping domain, researchers have used LLM Agents as virtual customers to test website features \cite{luUXAgentLLMAgentBased2025}, conduct automated A/B testing on design variants \cite{wangAgentAAutomatedScalable2025}, and evaluate agentic AI systems \cite{sunLLMAgentMeets2025}.
These applications build on the finding that LLM Agents are able to simulate ``believable'' human behavior \cite{parkGenerativeAgentsInteractive2023, parkGenerativeAgentSimulations2024}, whether those behaviors align with step-by-step human actions remains unclear.
Thus, a fundamental question remains unanswered: \textbf{How accurate can LLMs truly replicate human behavior?}

In this paper, we focus specifically on the human behavior simulation task that is to \textbf{generate the next action the user is most likely to perform in a multi-turn interaction session, based on the current observation and the history of past actions}. For instance, in an online shopping scenario, the model observes the current webpage context (e.g., a product list) and the user's action history (e.g., previous clicks or queries), and generates the next plausible action a human would take (e.g., add some product to the shopping cart).

Existing evaluations~\cite{parkGenerativeAgentsInteractive2023} of human behavior simulation primarily emphasize subjective measures of ``\textbf{believability}'' (``how much people feel it is like a human'') rather than the objective ``\textbf{accuracy}'' (``how much it acts like a human'').
The most relevant works that measure the objective model accuracy focus only on the final outcome of a task (e.g., purchasing the final product or not~\cite{yaoReActSynergizingReasoning2023},  or ultimately trusting the partner or not~\cite{xieCanLargeLanguage2024}), without examining whether the intermediate decision and action sequences align with those of actual humans.
Consequently, the field currently lacks a robust and quantitative understanding for assessing LLMs at the process-centric, action-level simulation of human behaviors.

To bridge this gap, we take online shopping as a case study and provide the first systematic evaluation of SOTA LLMs' accuracy in process-centric, action-level behavior simulation tasks.
We leverage a large-scale, real-world dataset consisting of 31,865 user click-through sessions from 3,526 users on an online shopping platform.
Each shopping session (Figure~\ref{fig:model-arch}) comprises a series of timestamp-aligned $\langle$context, action$\rangle$ pairs, where the context reflects the webpage observed by the user (e.g., product views, filter states), and the action denotes user inputs such as clicks, searches, or session termination actions.
In total, the dataset has 230,965 user actions, and the final outcomes of the sessions include 4,432 purchase actions and 27,433 session termination actions.
This dataset enables us to rigorously evaluate how accurately various LLMs can generate human-like behaviors at the action level.
We evaluate different models on the next action generation task, benchmarking both the accuracy of generated actions throughout the session and the F1 score for final session outcome prediction (i.e., purchase or not), following protocols similar to existing work.

Beyond evaluation, our dataset uniquely positions us to \textbf{fine-tune} LLMs to enhance their accuracy in behavior simulation tasks. 
While prior work has primarily relied on prompt-based approaches, we show that simply fine-tuning the model on user click-through data yields significantly better accuracy in action generation and session outcome prediction.
Furthermore, drawing inspiration from reasoning-augmented modeling \cite{deepseek-aiDeepSeekR1IncentivizingReasoning2025}, we hypothesize that exposing models to intermediate reasoning traces--even if synthetically generated--can enhance their ability to simulate human behavior.
To test this hypothesis, we augment our dataset with synthesized reasonings from action traces using Claude 3.5 Sonnet and fine-tune models using this augmented data (\texttt{$\langle$context, action, reasoning$\rangle$} triplets) to learn how to generate not only accurate actions but also the underlying reasoning.
Our results show that this reasoning-augmented fine-tuning further boosts model performance, highlighting the importance of modeling not just what humans do, but also why they do it.
Taken together, our results provide the first quantitative evidence that out-of-the-box LLMs cannot accurately simulate action-level human behavior in realistic settings, and our ablation and error analyses provide insights to support future research.

In summary, this paper has three main contributions:

\begin{itemize}
    \item We propose ShopCART, \textbf{the first quantitative and process-centric dataset} for evaluation of LLMs' ability to simulate human web action behaviors using real-world online shopping data\footnote{Code and data are available at \url{https://huggingface.co/datasets/NEU-HAI/ShopCART}}.

    \item We show that out-of-the-box LLMs cannot accurately predict human behavior in the next action prediction setting, while simple fine-tuning achieves substantially better performance.

    \item We demonstrate that \textbf{fine-tuning LLMs with synthesized reasoning traces further enhances their accuracy}, highlighting the importance of modeling not only what humans do but also why they do it for faithful human behavior simulation.

\end{itemize}

\section{Related Works}
\subsection{Simulation of Human Behavior with LLM }

The core function of the emerging LLM agent systems is their capability of generating human behaviors, in which a model takes a static user persona (e.g., preferences, demographics, or shopping habits), and the session data (e.g., a sequence of actions) as input to generate the next user action~\cite{lu2026agent,yao2026through}.
Such systems have been extensively utilized and tested to simulate human behavior in a variety of scenarios.
\citet{parkGenerativeAgentsInteractive2023} simulated social behavior using generative agents in a virtual town, producing ``believable'' interactions.  
\citet{xieCanLargeLanguage2024} studied LLM agents in Trust Games to assess their ability to model human trust behavior.  
\citet{parkGenerativeAgentSimulations2024} used LLMs to simulate responses from 1,052 individuals in a social science survey.  
To simulate UI interaction, \citet{luUXAgentLLMAgentBased2025} proposed UX\-Agent, enabling LLMs to operate within web environments for simulated usability testing.
Collectively, these studies underscore the growing potential of LLM-driven simulations to model and simulate complex, interpretable human behaviors.

However, the evaluation of these works remains limited in scope. Some focus on the subjective believability of process-centric action traces. 
For instance, \citet{luUXAgentLLMAgentBased2025} conducted qualitative interviews to assess participants' perceptions of the realism of their UXAgent system.
Similarly, \citet{parkGenerativeAgentsInteractive2023} proposed an evaluation framework that identified emergent social behaviors among generative agents.
On the other hand, works that pursue objective evaluation often do so in a single-shot, outcome-centric manner.
\citet{zhouWebArenaRealisticWeb2024} introduced WebArena, a controlled environment for benchmarking web agents based on task completion rates. ReAct \cite{yaoReActSynergizingReasoning2023} measured success rates in simulation environments, overlooking the accuracy of the model mimicking step-by-step human behavior.
To date, no prior work has focused on \textbf{objectively evaluating model-generated step-by-step actions}--that is, assessing whether a model's sequence of decisions faithfully aligns with human behavior at each step.

\subsection{Reasoning in Human Behavior Simulation}  

Building on the chain-of-thought prompting strategy \cite{weiChainofThoughtPromptingElicits2023}, numerous studies have incorporated reasoning mechanisms into human behavior simulation~\cite{yao2025dprf}.  
\citet{parkGenerativeAgentsInteractive2023} pioneered agents equipped with reflection modules that synthesize memory and social context to support introspective decision-making. 
ReAct \cite{yaoReActSynergizingReasoning2023} prompted models to generate reasoning traces and actions separately, improving task success rates in online shopping and gaming environments.  
\citet{gurRealWorldWebAgentPlanning2023} proposed WebAgent, which uses a dedicated reasoning model to plan sub-steps in web browsing tasks, enhancing control and planning in real-world browser simulations.  
Beyond single-agent reasoning, systems such as ChatDev \cite{qianChatDevCommunicativeAgents2024} and RepoAgent \cite{luoRepoAgentLLMPoweredOpenSource2024} adopt multi-agent setups, where agents with specialized roles (e.g., programmers, testers) engage in collaborative dialogues via structured prompts~\cite{chen2025multi}. These communicative exchanges support more robust collective reasoning, demonstrating how coordination between agents can improve the quality of generated reasoning traces.

However, the aforementioned works incorporate reasoning using prompt-only approaches for the action generation task. Whether reasoning can improve performance in \textit{fine-tuning} settings remains an open question.
Although datasets for reasoning behind model actions exist (e.g., for conversational agents~\citep{dongreReSpActHarmonizingReasoning2025}), there is currently no ready-to-use dataset specifically designed for human behavior simulation that includes both human actions and the corresponding reasoning behind those actions.
To address a similar cold-start problem in reinforcement learning, \citet{deepseek-aiDeepSeekR1IncentivizingReasoning2025} constructed a small-scale dataset of long reasoning traces by synthesizing examples through few-shot prompting, followed by reflection and verification. Inspired by their methodology, we apply a similar strategy to synthesize reasoning traces in the online shopping domain. This enables us to investigate whether integrating reasoning into fine-tuning can enhance a model's ability to accurately simulate human behavior.

\section{Method}

\subsection{Task Definition}  

In this section, we formally define the proposed human behavior simulation task: in the online shopping scenario, a shopping session is represented as a sequence of user actions $a_{1 \dots t \dots N}$, always starting with a \texttt{search action} and concluding with either a \texttt{product purchase action} or a \texttt{termination action} (i.e., the user closing the browser window).  

At each time step \( t \), the model is tasked with generating both \textbf{the reasoning \( r_t \)} and \textbf{the action \( a_t \)}. The model input includes the current context \( c_t \) (what the user currently observes), a sequence of previous contexts (what the user has observed) \( c_{1 \dots t-1} \), a sequence of previous actions \( a_{1 \dots t-1} \) (what the user has done), and the corresponding synthetic reasoning trace \( r_{1 \dots t-1} \) (why the user did that action) within the same session. Formally, the model learns a function \( f \) such that:  
\[
    f(c_{1 \dots t}, a_{1 \dots t-1}, r_{1 \dots t-1}) = r_t, a_t
\]

\paragraph{Observation Context}

The \textbf{Context} (or ``observation space'') of the web agent encompasses all available information on a webpage, including textual content, metadata, visual elements, and structural data. This context is designed to reflect how a human perceives and interprets a webpage, allowing the agent to process relevant features and perform tasks such as navigation, information retrieval, and interaction with page elements.  

Previous research has explored various context formats. Some methods rely on manually structured information parsing \cite{yaoWebShopScalableRealWorld2022}, while others utilize raw HTML representations \cite{gurRealWorldWebAgentPlanning2023} or accessibility tree embeddings \cite{zhouWebArenaRealisticWeb2024}. However, both approaches have notable limitations: manually structured information parsing requires significant human effort to develop parsing rules \cite{yaoWebShopScalableRealWorld2022}, while raw HTML representations often contain extraneous information (e.g., JavaScripts) that is not observable to human users, which may bias the LLM performance.  

To ensure the adaptability to unseen websites, we define and implement a \textbf{simplified HTML} format as our context representation. This format removes non-relevant elements such as scripts, CSS, and purely visual components while preserving essential structural information. Using simplified HTML offers several advantages compared to custom-defined formats (Domain-Specific Languages, DSLs) or markdown-based representations: (1) important structural elements, such as lists and tables, remain intact, and (2) LLMs are already familiar with the HTML format, eliminating the need to redefine common elements like ``button'' and ``input''.

The LLM agent needs to refer to specific elements within the HTML, such as identifying the exact button it intends to click. Since there is no built-in method to uniquely identify HTML elements, prior work has proposed approaches like assigning sequential IDs to elements \cite{kohVisualWebArenaEvaluatingMultimodal2024} or manually defining descriptive names for elements, such as \code{searchbox} \cite{yaoWebShopScalableRealWorld2022}. Following prior work \cite{luUXAgentLLMAgentBased2025}, we assign a unique hierarchical \textit{name} in natural language to each interactable element, including links, buttons, and input fields. This name is constructed by incorporating the names of all parent nodes. For instance, if a \code{<a>} tag named \code{view_product} resides within a \code{<div>} named \code{columbia_shirt}, the resulting hierarchical name will be \code{columbia_shirt.view_product}.

\paragraph{Reasoning}

Reasoning trace refers to a natural language description that articulates the reasoning and explanation behind an action.
For example, if the context is a search results page displaying a list of clothes, and the generated action is clicking on the \code{"4 stars and up"} product review filter, the generated reasoning might be: \textit{``I want to find a comfortable piece of clothing, so I'm looking for options with high ratings.''}
In our domain, the reasoning trace is missing from any real-world online shopping data, so we use LLM to synthesize it (Section~\ref{sec:data-synthesize}).
The generated reasoning provides insight into the model's thinking process, enhancing the transparency of the model's generations.

\paragraph{Action}
Previous research has explored various approaches to defining action spaces, including task-specific semantic actions such as ``searching'', ``adding items to a cart'', and ``making purchases'' \cite{yaoWebShopScalableRealWorld2022}, as well as browser-level interactions like \texttt{typing} and \texttt{clicking} \cite{luUXAgentLLMAgentBased2025}. 

To ensure the adaptability of our framework beyond online shopping tasks, we define the action space at the level of \textbf{raw browser actions}, rather than at the level of task-specific semantics.
The action space of our model consists of three fundamental browser operations: \code{click}, \code{type_and_submit}, and \code{terminate}.
This abstraction allows the system to generalize across different environments while maintaining task flexibility.
\label{sec:action-definition}

\subsection{Synthesized Reasoning Trace}
\label{sec:data-synthesize}
Reasoning traces are crucial for understanding users' action choices but are difficult to collect; thus, they are often not readily available.
We employ a reasoning synthesis pipeline to generate them using an LLM.
To guide the reasoning generation process, we provide the LLM with the observation context and the corresponding action.
Additionally, we record real human customers' think-aloud shopping sessions \cite{ecclesThinkAloudMethod2017} as in-context learning examples.
We then prompt the LLM to generate a free-text reasoning explaining the user's decision.
Following recent works on reasoning-augmented LLMs~\cite{weiChainofThoughtPromptingElicits2023,deepseek-aiDeepSeekR1IncentivizingReasoning2025}, \textbf{the synthesized reasoning is not intended to replicate the actual human thought process}. Instead, its purpose is to enhance the model's predictive accuracy by providing structured intermediate representations that help the LLM better align actions with contextual cues.
This approach ensures that the reasoning traces are coherent with the observed actions while improving the model's behavioral fidelity and explainability.

\subsection{Model Architecture}

To incorporate these enriched action traces, we build on existing pre-trained LLMs as our base models.
The input to the model consists of two components: \textbf{(1) a sequence of historical contexts} (what the user observed), and \textbf{(2) the corresponding actions and generated reasoning trace} (what the user did and why).

During the \textbf{training stage}, the model receives the full sequence of a user session—including context, synthetic reasoning, and action—as a single concatenated input. 
The training objective is to minimize the next-token prediction loss for the reasoning and action tokens, while the loss for the context tokens is masked out. 
Subsequently, in the \textbf{evaluation stage}, the model is provided with historical context and past reasoning traces and actions, and is asked to first generate the reasoning for the next action; then, based on the generated reasoning, it generates the next action.

The model generates both reasoning and action in sequence using a multi-turn conversation format. Each action in the dataset is represented as a two-turn interaction. In the first turn, the model is prompted with the observation context to generate the next reasoning. Then we provide a hard-coded message (\texttt{<|end\_of\_rationale|>}) to prompt the model to produce the corresponding action.

\section{Experiments}
\subsection{Dataset Construction}

Our dataset, \datasetname{}\footnote{\url{https://huggingface.co/datasets/NEU-HAI/ShopCART}}, was constructed using data from Amazon.com.
The dataset contains 31,865 user sessions from 3,526 users in the online shopping scenario, comprising 230,965 user actions.
The session's final outcome includes 4,432 purchase actions and 27,433 session termination actions.
We leveraged our data synthesis pipeline, detailed in Section \ref{sec:data-synthesize}, to generate the synthetic reasoning for each action based on the context using Claude-3.5-Sonnet.
\textbf{The dataset was derived from traffic logs of a small group of users who explicitly opted into a beta testing feature and consented to data collection during the process.
Users who opted out of data collection were excluded from the dataset.
The data was processed with an LLM to remove any personally identifiable information.
}

Additionally, to test whether our conclusion can be generalized to other datasets, we also repeat our setup on the OPeRA dataset \cite{wangOPeRADatasetObservation2025}. OPeRA is a dataset of Observation, Persona, Rationale, and Action collected from 51 real users across 692 online shopping sessions, providing 28,904 time-aligned ⟨observation, action⟩ pairs together with 604 human-annotated rationales and detailed persona profiles.

We extracted pairs of user actions and context from the cleaned data. These raw data were then structured into the standardized format defined in Section \ref{sec:action-definition}.

\begin{table*}[t]
\centering
\small

\begin{booktabs}{
    width=\linewidth,
    colspec={lcccccc},
    cell{1}{2,5} = {c=3}{},
    cell{1}{1} = {r=2}{},
    column{1}={font=\small},
    row{1,2}={font=\bfseries},
    cell{3,15,22}{1} = {c=7}{bg=black!30!white,font=\bfseries},
    row{3,15,22}={rowsep=4pt},
    cell{24,26,28}{1} = {r},
}
\toprule
Model                & {Generated Next Action} & & & {Session Outcome} \\
\cmidrule[r]{2-4} \cmidrule[l]{5-7}
                     & Accuracy                & $\%\Delta$ vs Base & v.s. DS-R1 & F1 Score             & $\%\Delta$ vs Base & v.s. DS-R1 \\
\midrule
Open-Source Models   &                         &                    &                  &                      &                    &                  \\
DeepSeek-R1          & 11.86\%                 & -                  & -                & 20.01\%              & -                  & -                \\
Llama 3.1 8B         & 5.05\%                  & -                  & \highlightRelevantValue{-6.81} & 10.87\%              & -                  & \highlightRelevantValue{-9.14} \\
Llama 3.1 70B        & 8.19\%                  & -                  & \highlightRelevantValue{-3.67} & 12.69\%              & -                  & \highlightRelevantValue{-7.32} \\
Mixtral 8x7B         & 5.41\%                  & -                  & \highlightRelevantValue{-6.45} & 13.16\%              & -                  & \highlightRelevantValue{-6.85} \\
Qwen2.5-70B          & 6.46\%                  & -                  & \highlightRelevantValue{-5.40} & 11.96\%              & -                  & \highlightRelevantValue{-8.05} \\
Qwen2.5-7B           & 4.25\%                  & -                  & \highlightRelevantValue{-7.61} & 11.94\%              & -                  & \highlightRelevantValue{-8.07} \\
Qwen2.5-3B           & 3.91\%                  & -                  & \highlightRelevantValue{-7.95} & 10.87\%              & -                  & \highlightRelevantValue{-9.14} \\
Qwen2.5-1.5B         & 3.27\%                  & -                  & \highlightRelevantValue{-8.59} & 7.94\%               & -                  & \highlightRelevantValue{-12.07} \\
Mistral-7B-v0.3      & 4.25\%                  & -                  & \highlightRelevantValue{-7.61} & 11.27\%              & -                  & \highlightRelevantValue{-8.74} \\
Llama 3.2 3B         & 2.93\%                  & -                  & \highlightRelevantValue{-8.93} & 8.60\%               & -                  & \highlightRelevantValue{-11.41} \\
Llama 3.2 1B         & 3.71\%                  & -                  & \highlightRelevantValue{-8.15} & 3.09\%               & -                  & \highlightRelevantValue{-16.92} \\
Proprietary Models   &                         &                    &                  &                      &                    &                  \\
Claude 3.5 Haiku     & 9.18\%                  & -                  & \highlightRelevantValue{-2.68} & 14.77\%              & -                  & \highlightRelevantValue{-5.24} \\
Claude 3 Opus        & 6.78\%                  & -                  & \highlightRelevantValue{-5.08} & 15.08\%              & -                  & \highlightRelevantValue{-4.93} \\
Claude 3 Sonnet      & 8.42\%                  & -                  & \highlightRelevantValue{-3.44} & 17.40\%              & -                  & \highlightRelevantValue{-2.61} \\
Claude 3.5 Sonnet    & 9.72\%                  & -                  & \highlightRelevantValue{-2.14} & 15.91\%              & -                  & \highlightRelevantValue{-4.10} \\
Claude 3.5 Sonnet v2 & 11.69\%                 & -                  & \highlightRelevantValue{-0.17} & 18.54\%              & -                  & \highlightRelevantValue{-1.47} \\
Claude 3.7 Sonnet    & 9.34\%                  & -                  & \highlightRelevantValue{-2.52} & 12.81\%              & -                  & \highlightRelevantValue{-7.20} \\
Fine-tuned Models    &                         &                    &                  &                      &                    &                  \\
Qwen2.5-7B~          & \underline{16.67\%}                 & -                  & \highlightRelevantValue{4.81}  & 26.92\%              & -                  & \highlightRelevantValue{6.91}  \\
+ reasoning          & \textbf{17.26\%}                 & \highlightRelevantValue{3.54}  & \highlightRelevantValue{5.40}  & \underline{ 33.86\%}              & \highlightRelevantValue{25.78} & \highlightRelevantValue{13.85} \\
Mistral-7B-v0.3~     & 14.17\%                 & -                  & \highlightRelevantValue{2.31}  & 17.99\%              & -                  & \highlightRelevantValue{-2.02} \\
+ reasoning          & 15.84\%                 & \highlightRelevantValue{11.79}  & \highlightRelevantValue{3.98}  & 30.12\%              & \highlightRelevantValue{67.43}  & \highlightRelevantValue{10.11} \\
Llama-3.2-3B~        & 9.31\%                  & -                  & \highlightRelevantValue{-2.55} & 4.73\%               & -                  & \highlightRelevantValue{-15.28} \\
+ reasoning          & 15.77\%                 & \highlightRelevantValue{69.39}  & \highlightRelevantValue{3.91}  & \textbf{33.99\%}              & \highlightRelevantValue{618.60} & \highlightRelevantValue{13.98} \\
\bottomrule
\end{booktabs}
\caption{Model performance. The table shows model accuracy in two tasks: the process-centric \textbf{action generation} task and the \textbf{outcome}-centric final purchase prediction task of the session. More models' performances are in the Appendix. DS-R1: performance comparison with DeepSeek-R1.}
\label{tab:results}
\end{table*}

\begin{table}[t]
\centering
\begin{booktabs}{
  colspec = {lcc},
  cell{10}{1}={r},
  cell{2,8}{1} = {c=3}{bg=black!30!white,font=\bfseries},
}
\toprule
Model & Action Gen. Acc & Session F1 \\
\midrule
Pretrained LLMs &  &  \\
GPT-4.1 & 21.28\% & 51.17\% \\
DeepSeek-R1 & 15.74\% & 47.92\% \\
Claude-3.7 & 10.08\% & 43.10\% \\
Llama-3.3-70B & 8.76\% & 34.19\% \\
Qwen-2.5-7B & 4.10\% & 41.11\% \\
Fine-tuned LLMs &  &  \\
Qwen-2.5-7B & 32.04\% & 71.38\% \\
 + reasoning & \textbf{35.14\%} & \textbf{75.85\%} \\
 \bottomrule
\end{booktabs}
\caption{Model performance on the OPeRA dataset.}
\label{tab:opera}
\vspace{-\baselineskip}
\end{table}

\subsection{Evaluation and Metrics}

\paragraph{Evaluation Dataset}  

We used a subset of the dataset that was not used during training as the test set, ensuring that no user sessions in the test set were seen by the model during fine-tuning.
To create test cases, we took the second and all subsequent actions within each session, excluding the first action, since it lacks any preceding context.
For each test case, the model was provided with the historical context, along with all previous actions and corresponding synthetic reasoning traces in the same session.
The model is \textbf{first tasked with predicting the next reasoning, and then, based on the reasoning it generated}, it produces the corresponding action.

\paragraph{Prompt-Only Baseline}
To assess pre-trained models' ability in predicting human behavior, we evaluated a set of instruction-tuned LLMs under the in-context learning (ICL) setting (a.k.a., prompt-based setting). Specifically, we used several variants of Claude, LLaMA, and Mistral as representatives for general-purpose pre-trained LLMs, along with DeepSeek-R1 as a representative for reasoning LLMs.
To preserve user privacy, we run open-source models on our own GPU cluster and access proprietary models through their official APIs.

In this setup, each model was provided with the historical context and previous user actions from the session, and prompted to generate both the next reasoning and the next action.
The generated actions were used to compute macro accuracy for evaluation. 
These baselines reflect the commonly adopted approach of using powerful LLMs without domain-specific fine-tuning, allowing us to directly assess the impact of fine-tuning on realistic human behavior simulation.

\paragraph{Evaluation Metrics}  

We evaluate model performance across two key dimensions: \textbf{Next Action Generation} and \textbf{Session Outcome Classification}.

For \textbf{Next Action Generation}, we evaluate whether the model can accurately generate user actions during the process.
We use an \textit{\textbf{exact match accuracy}}, where a generation is considered correct only if the action type, action target (e.g., search box or product link), and action attribute (such as search keyword) exactly match the ground truth. 
To avoid skewing the results toward longer sessions, we compute per-session accuracy first and then average across sessions, ensuring each session contributes equally to the final score.

To evaluate models' performance beyond human behavior simulation to shopping prediction, we introduce an additional task and metric focused on the \textbf{session final outcome}. 
The evaluation setting remains the same: the model is given the ground-truth session history up to the last step and tasked with generating the next (and final) action. 
Since the final action is always either a \code{click} action on the \code{buy now} button or a \code{terminate} action indicating closing the browser window, we evaluate the binary classification performance using the F1 score. 
This allows us to assess how well the model distinguishes between these two critical session outcomes (\texttt{buy} or \texttt{termination}).

\begin{figure*}[t]
    \centering
    \includegraphics[width=.9\linewidth]{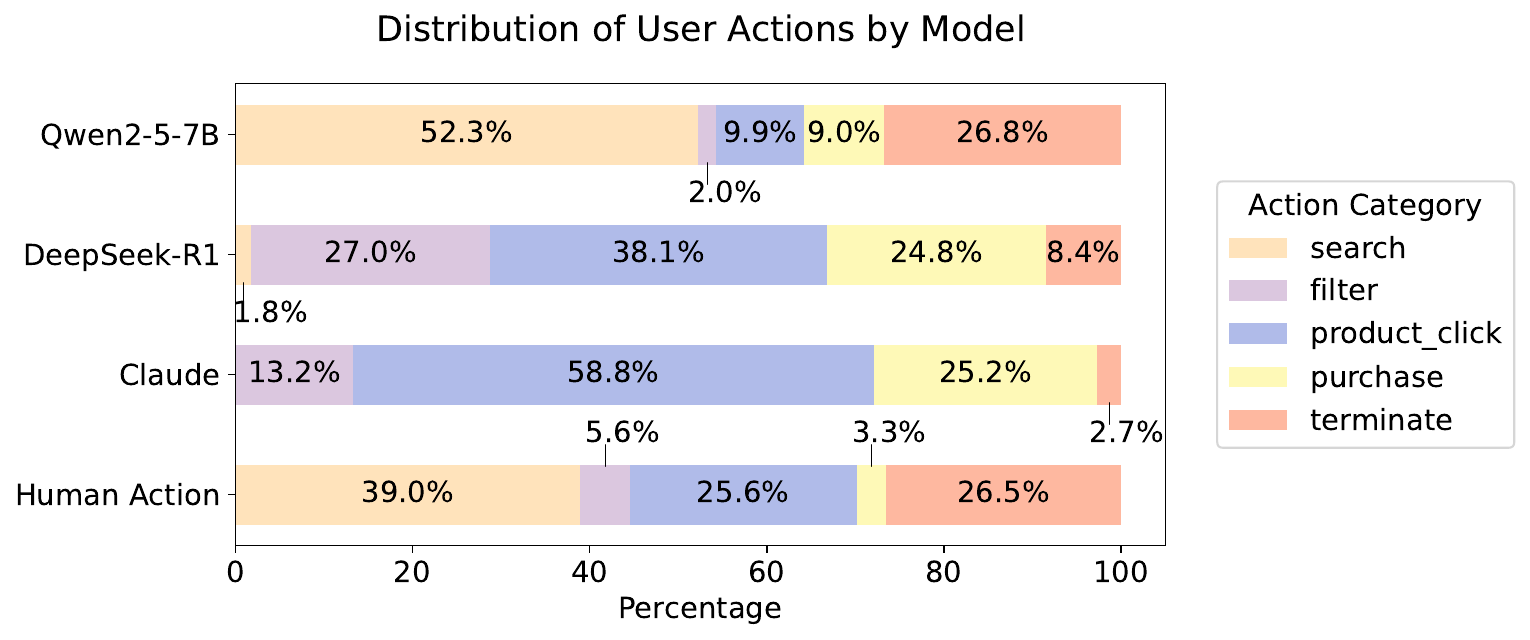}
    \caption{Action categories of human groundtruth, and generated by prompt-based Claude and by our fine-tuned models.}
    \label{fig:action-distribution}
    \vspace{-1\baselineskip}
\end{figure*}

\subsection{Experimental Setup}
We fine-tuned multiple language models to evaluate their performance on the action generation task. The models used in our experiments include:
\begin{itemize}
    \item \textbf{Fine-Tuned Models:} Different versions of Llama 3.2, Qwen 2.5, and Mistral.
    \item \textbf{Baseline Models:} Different versions of Claude, Llama, Mistral, and DeepSeep.
\end{itemize}
All fine-tuned models were trained using the same dataset and pipeline to ensure a fair comparison. 

Model training was performed on a GPU cluster consisting of NVIDIA H200 GPUs, with each training job utilizing eight nodes $\times$ eight GPUs, for a total of 64 GPUs, each with 140 GB of GPU memory. A typical job takes 3 hours on 64 GPUS, and in total, we used about 3700 H200 GPU hours for our experiments.

We employed Fully Sharded Data Parallel \cite{zhaoPyTorchFSDPExperiences2023} for efficient training. All sequences were padded or truncated to a context length of 40k tokens. We used a per-device batch size of 1, resulting in a global batch size of 64. The learning rate was set to 2e-5 with a cosine scheduler for adaptive learning rate adjustment. Models were trained for 1 epoch. For example, training Mistral-7B-v0.3 with a 40k token context window in our setup requires approximately 130GB of GPU memory per GPU, which is the largest model we can train.

\begin{figure*}
    \centering
    \includegraphics[width=.95\linewidth]{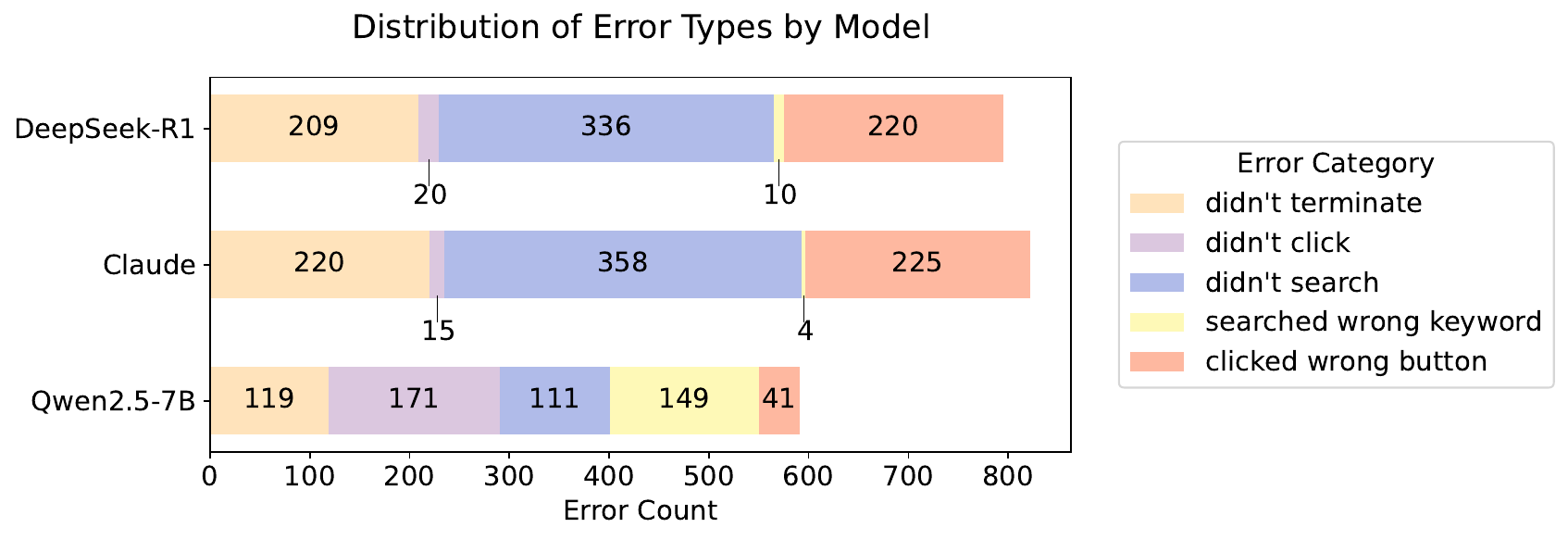}
    \caption{Error Type Analysis of different models.}
    \label{fig:error-analysis}
    \vspace{-1\baselineskip}
\end{figure*}

\subsection{Evaluation Results and Analysis}

Following previous works \cite{lutzWILBURAdaptiveInContext2024, dengMind2WebGeneralistAgent2023}, we evaluate the performance of prompt-based LLMs in simulating human behavior on shopping scenarios, with results presented in Table~\ref{tab:results}.
Our findings indicate that while state-of-the-art LLMs have demonstrated strong capabilities in various tasks and domains, \textit{their ability to accurately simulate human behaviors remains limited}, with Claude 3.5 sonnet v2 achieving 11.69\% accuracy in next action prediction.
Compared to general instruction-tuned models, reasoning-focused DeepSeek-R1 achieved a higher accuracy of \textbf{11.86\%} in the action generation task, suggesting that incorporating reasoning mechanisms offers some advantage.
Similarly, for final outcome prediction, DeepSeek-R1 achieved an F1 score of \textbf{20.01\%}, outperforming other baseline models.
Additionally, larger models usually perform better than their smaller variants.
These results indicate that reasoning ability positively impacts performance on human behavior prediction tasks, supporting the hypothesis that such tasks benefit from models trained with reasoning-oriented objectives.

We then fine-tuned various smaller open-sourced LLMs, including LLaMA 3.2, Qwen 2.5, and Mistral, using our training dataset. The results demonstrate that fine-tuning LLMs with action traces and synthesized reasoning traces significantly enhances performance. \textbf{Qwen 2.5-7B achieved \textbf{17.26\%} accuracy in action generation, significantly surpassing DeepSeek-R1 by \textbf{5.4\%} {($p < 10^{-10}$, McNemar's test)}.} Similarly, LLaMA 3.2-3B reached an F1 score of \textbf{33.99\%} on the final outcome prediction task, further confirming the effectiveness of fine-tuning.
Additionally, all fine-tuned models significantly outperformed their own ICL variants ($p < 10^{-5}$, McNemar's test).
These findings underscore that incorporating domain-specific fine-tuning with synthesized reasoning traces leads to substantial improvements in the accuracy of human online shopping behavior simulation.

Additionally, a similar trend is observed on the OPeRA dataset. Table~\ref{tab:opera} reports model performance in terms of Action Generation Accuracy and Session Outcome F1. Among pretrained models, GPT-4.1 achieves the strongest results, followed by DeepSeek-R1 and Claude-3.7, while smaller open-weight models lag behind. In contrast, fine-tuned Qwen models substantially outperform all pretrained baselines while being much smaller, with further gains obtained by incorporating reasoning signals. 
These results further indicate that the ability of pretrained LLMs to accurately simulate human behavior remains limited, and task-specific fine-tuning is critical for achieving faithful behavior modeling.

Figure \ref{fig:action-distribution} compares the distribution of actions generated by different models with real human behavior. Human users rarely apply filters and instead rely heavily on iterative search, averaging 2.82 searches per session, which is over seven times more frequent than filter actions. This reflects natural behaviors such as correcting typos and revising keywords (as shown in Sec.~\ref{sec:error-anaysis}). In contrast, pre-trained LLMs such as Claude 3.5 Sonnet and DeepSeek-R1 tend to stick to the initial search keyword without revision, overuse filtering actions, and produce disproportionately high purchase rates that diverge from actual user behavior. 
We hypothesize that this bias arises because \textbf{existing LLM agent benchmarks like WebShop\cite{yaoWebShopScalableRealWorld2022} and WebArena\cite{zhouWebArenaRealisticWeb2024} primarily evaluate task completion} (i.e., making a purchase), which incentivizes models to optimize for purchase-heavy trajectories rather than realistic browsing and search behavior.

The fine-tuned models exhibit a distribution that more closely aligns with human action patterns, capturing a more natural balance between search refinement, product clicks, and minimal reliance on filtering. Unlike pre-trained models, they do not prematurely converge on purchase-oriented behavior and instead reflect the exploratory and iterative nature of real user sessions.

\begin{table*}[t]
    \centering
    \small
    \begin{booktabs}{lll}
    \toprule
     & \textbf{Example 1} & \textbf{Example 2} \\
    \midrule
Previous Action & search for ``disney gift'' & search for ``tee conector'' \\
Human Next Action & search for ``disney gift card'' & search for ``tee connector'' \\
Qwen-2.5-7B & search for ``disney gift card'' & search for ``tee connector'' \\
Claude & click on \code{disney_gift_card_...} & click on \code{spalolen_30_pack_...} \\
    \bottomrule
    \end{booktabs}
    \caption{Model predictions vs. human next actions.}
    \label{tab:model-examples}
    \vspace{-\baselineskip}
\end{table*}

\subsection{Ablation Study}  

To evaluate the impact of synthesized reasoning traces on the model's action generation capability, we conducted an ablation experiment. In the base setting, we removed reasoning traces from both the training and evaluation stages to isolate the contribution of reasoning-augmented learning. This setup allows us to directly assess whether exposure to synthesized reasoning traces improves the model's ability to generate user actions and predict outcomes.

From Table~\ref{tab:results}, most models exhibited substantial improvements in action generation accuracy when trained with reasoning traces, with relative gains ranging from 3.54\% to 69.39\%. Similar benefits were observed for final outcome prediction, where F1 scores increased by 25.78\% to over 600\% across models. For example, Qwen2.5-7B achieved a \textbf{33.86\%} F1 score with reasoning traces but dropped to \textbf{26.92\%} without them, underscoring the importance of explicit reasoning guidance. These results confirm that incorporating synthesized reasoning traces enhances both step-wise action generation and final outcome modeling, reinforcing their value in fine-tuning models for human behavior simulation.

\subsection{Error Analysis}
\label{sec:error-anaysis}
We analyze model behavior across different error types, comparing two ICL models, Claude, representing general-purpose chat models, and DeepSeek-R1, representing reasoning-augmented models, with a fine-tuned compact model, Qwen2.5–7B.
We focus on five distinct error types. \textbf{Didn't terminate}, \textbf{didn't click}, and \textbf{didn't search} indicate cases where the model chose a different action instead of terminating, clicking, or searching, as the human user did. \textbf{Searched wrong keyword} refers to instances where the model performed a search like the user, but used a different (and incorrect) search query. \textbf{Clicked wrong button} captures cases where the model clicked, but on a different button than the one selected by the human user.
Illegal actions generated by models are excluded from this analysis.
Overall, Claude and DeepSeek-R1 exhibit very similar error profiles, suggesting that the reinforcement learning process used to incorporate reasoning introduces similar biases to those seen in standard chat-based ICL models. A major shared failure mode of ICL models is their tendency to continue sessions or make the purchase even when human users would have terminated them by closing the browser window. This aligns with earlier findings that ICL agents are more likely to complete purchases, likely because current LLMs are optimized to fulfill task goals rather than follow subtle social cues or termination heuristics. In contrast, fientuned Qwen2.5–7B demonstrates more accurate alignment with human termination behavior.

We also observe that Qwen-2.5–7B better captures iterative search behaviors commonly exhibited by real users, such as retrying a query with corrected keywords or fixing typos after an unsatisfactory result. In contrast, ICL-based models like Claude tend to persist with the original query result and proceed to a different type of action, rather than issuing a refined search.
As shown in Table~\ref{tab:model-examples}, Qwen-2.5–7B more closely matches human revisions in both examples, whereas Claude chooses to visit the product shown on the current search result page.
These findings suggest that fine-tuning not only reduces the overall error rate but also enhances the model's ability to capture fine-grained user behavior patterns, such as correcting typos, retrying failed searches, and deciding when to terminate a session.

\section{Discussion and Future Works}  

\subsection{Action Misalignment Between Human and Large Language Models}

Existing research has shown that LLMs can generate highly ``believable'' human behavior simulations, supporting various interactive and social simulation scenarios.  
However, our results indicate that general-purpose pretrained LLMs, despite their strong ability in a variety of tasks and applications, struggle to \textbf{accurately} generate user actions.
This is reflected in their performance—DeepSeek-R1 and Claude 3.5 Sonnet v2, for example, achieved only \textbf{11.86\%} and \textbf{11.69\%} accuracy, respectively, on the next action generation task. These findings suggest that while LLMs may appear human-like and produce plausible and believable behaviors, accurate step-wise next action prediction requires additional fine-tuning and explicit alignment with real human actions.  

Building on this observation, we find that current LLM agents are often misaligned with real human users, particularly in the online shopping domain. For example, as shown in Figure~\ref{fig:action-distribution} and Figure~\ref{fig:error-analysis}, out-of-the-box LLMs tend to make more purchases than human users.
We hypothesize that this stems from the training objectives of many LLMs, which are typically optimized for the final task completion (e.g., making purchases) and evaluated based on task completion and task efficiency (i.e., completing the task with a minimal number of steps).
To enable accurate human behavior simulation, it is essential to close this gap and ensure models capture what a real user will do under certain circumstances.

Part of this accuracy gap may also stem from the mismatch between the information available to the model and the information available to real users. In our setup, models receive simplified HTML as input, whereas human shoppers make decisions based on rendered visual layouts, product images, and other perceptual cues that are absent from the markup. This means that even a perfect text-based model would lack signals that are central to human decision-making, such as the visual appeal of a product thumbnail or the spatial layout of a page. Closing this modality gap, for instance by grounding agents in rendered screenshots via vision-language models, is a necessary step toward improving absolute prediction accuracy.

It is worth noting that exact-match next-action prediction is an inherently difficult task and a strict metric: the model must produce the precise action a specific user took, including the exact search query, the exact product clicked, and the exact option selected. Our results show that current LLMs cannot yet achieve this level of step-wise accuracy. Nevertheless, fine-tuned models do exhibit qualitative similarity to real user behavior: they learn to browse, compare, revise search queries, and terminate sessions in patterns that resemble human trajectories (Section~\ref{sec:error-anaysis}). This suggests that LLM agents may already be useful for applications that depend on aggregate behavioral plausibility, such as usability testing or traffic simulation, but are not yet reliable enough for applications that require precise individual-level fidelity, such as A/B testing, where small differences in action distributions can lead to incorrect conclusions.

\textbf{We also advocate for the development of evaluation metrics that reflect the irrational nature of human behavior.}
On one hand, many user actions are rational and goal-directed, where a single correct next step exists. On the other hand, users also exhibit irrational or weakly deterministic behaviors, such as semi-random clicks, and typos that even the same person may not replicate. 
A robust metric should not penalize plausible variants that remain consistent with human trajectories and should explicitly assess an agent's ability to reproduce human-like error patterns.
Possible directions include partial-match scoring that credits semantically equivalent actions (e.g., two search queries with the same intent), action-type-weighted metrics that assign higher importance to critical decision points such as purchases over incidental clicks, and distributional measures that compare aggregate action statistics between simulated and real sessions rather than requiring exact per-step alignment.

\subsection{Reasoning in Next Action Generation}

Our experiments further highlight the critical role of synthesized reasoning traces in improving action generation. 
Removing reasoning traces from the training data led to a notable performance drop on most models. 
These results suggest that reasoning traces not only enhance model interpretability but also serve as a guiding mechanism, enabling the model to make more contextually appropriate and human-aligned decisions. 

\subsection{Reasoning and Human Cognition}

There are two distinct notions of reasoning in LLM-based behavior simulation. The first is \textbf{accuracy-oriented reasoning}, where effective reasoning is defined as reasoning that leads to better downstream task performance \cite{deepseek-aiDeepSeekR1IncentivizingReasoning2025,weiChainofThoughtPromptingElicits2023}. The second is \textbf{cognition-oriented reasoning}, where the goal is to faithfully model the actual cognitive processes of human users.
Our work, along with most prior work on reasoning-augmented LLMs, adopts the first definition: the synthesized reasoning traces are generated by an LLM and optimized to improve action prediction accuracy, not to replicate how humans actually think. For applications that depend on behavioral fidelity, such as predicting what a user will do next, this is sufficient. However, for applications that require understanding \emph{why} a user acts a certain way, such as simulated user interviews or cognitive modeling, the second definition becomes necessary, and training on authentic human think-aloud data would be more appropriate.

\subsection{Future Works}

Future research should move beyond treating reasoning traces as static supervision and instead focus on actively enhancing the model's reasoning abilities. 
One direction is to \textbf{use reinforcement learning to improve the reasoning for action prediction accuracy}, where better reasoning is defined by its ability to produce more accurate action prediction. 
Another direction is to \textbf{train models to generate reasoning that more closely reflects human cognitive processes}, so the reasoning itself can serve as a source of qualitative insight when analyzing or interacting with LLM agents.
It would also be valuable to evaluate alternative reasoning trace generators beyond Claude-3.5-Sonnet, such as GPT-4 or open-source models, to assess how the choice of generator affects downstream performance.
Additionally, we adopted a multi-turn conversation format where reasoning and action are generated in separate turns, which means the model fine-tuned with reasoning traces cannot be straightforwardly evaluated in action-only mode. Future work could explore using standardized reasoning tokens (e.g., \texttt{<think>...</think>}) to unify the training and inference protocols, enabling direct comparison between reasoning-augmented and action-only settings within a single model.
Finally, our text-only setup using simplified HTML does not capture the visual cues (e.g., product images, layout saliency) that real users rely on. Incorporating vision-language models (VLMs) to process rendered web pages could better reflect actual human decision-making inputs and potentially improve behavioral fidelity.

\section{Conclusion}

In this work, we present the first quantitative, process-centric evaluation of state-of-the-art LLMs for simulating human behavior in an online shopping task.
Our study demonstrates that LLMs fine-tuned with real-world human behavioral data and synthesized reasoning traces showed a significant enhancement in their ability to generate user actions across different datasets.
These results underscore the critical role of explicit reasoning traces in aligning model predictions with human behavior.
By enriching behavioral datasets with structured reasoning, we move closer to accurate and interpretable simulations of human behavior.
Our findings highlight a promising direction for the development of LLM agent systems capable of producing realistic and explainable human-like behaviors in online shopping and other interactive domains.

\section*{Limitations}

Our study has several limitations that should be considered when interpreting the results. 
First, following recent works \cite{deepseek-aiDeepSeekR1IncentivizingReasoning2025}, we only evaluated the extent to which synthetic reasoning traces improve model performance.
Conducting human evaluations on the interpretability and usefulness of the generated reasoning traces could better assess how well these traces support human comprehension and trust in the model's predictions.
Second, we have not yet evaluated the model on real human-annotated datasets containing authentic reasoning traces, making it unclear how well the synthesized reasoning trace aligns with human reasoning. Additionally, the process of generating synthesized reasoning traces may introduce unintended biases, potentially impacting prediction accuracy and interoperability. Generating synthesized data with a proprietary model making the cost of reproducing the result high.
Finally, to simplify the experimental setup, we limited the action space to basic browser operations such as \code{type} and \code{click}. Incorporating more complex interactions, such as scrolling, waiting, or hover actions, would allow for a more realistic simulation of human behavior in web environments and offer deeper insights into how LLMs handle nuanced browser-based simulation.
Additionally, our evaluation relies on simplified HTML as model input, whereas real users make decisions based on rendered visual layouts and product images rather than raw markup. We adopt this text-only design following standard practice in web agent research \cite{dengMind2WebGeneralistAgent2023,zhouWebArenaRealisticWeb2024,lutzWILBURAdaptiveInContext2024,qiWebRLTrainingLLM2025}, and note that current VLMs still underperform text-based LLMs on web interaction tasks (e.g., 66.3\% success rate on WebArena vs.\ 38.35\% on VisualWebArena \cite{kohVisualWebArenaEvaluatingMultimodal2024}).
Future work should explore incorporating VLMs to process rendered web pages and improve behavioral fidelity.
Furthermore, our study uses online shopping as a single case study. While shopping represents a complex, multi-turn decision-making process, it remains an open question whether our findings generalize to other interactive domains such as travel booking, customer support, or information seeking. Evaluating cross-domain transferability is an important direction for future work.

\newpage

\bibliography{custom}
\clearpage

\appendix

\begin{table*}
\centering
\begin{booktabs}{
    width=\linewidth,
    colspec={lcccccc},
    cell{1}{2,5} = {c=3}{},
    cell{1}{1} = {r=2}{},
    column{1}={font=\small},
    row{1,2}={font=\bfseries},
    cell{3,14,21}{1} = {c=7}{bg=black!30!white,font=\bfseries},
    cell{23,25,27,29,31,33}{1} = {r},
}
\toprule
Model                & {Generated Action} & & & {Session Outcome} \\
\cmidrule[r]{2-4} \cmidrule[l]{5-7}
                     & Accuracy                & $\%\Delta$ vs Base & v.s. DS-R1 & F1 Score             & $\%\Delta$ vs Base & v.s. DS-R1 \\
\midrule
Open-Source Models   &                         &                    &                  &                      &                    &                  \\
DeepSeek-R1          & 11.86\%                 & -                  & -                & 20.01\%              & -                  & -                \\
Llama 3.1 8B         & 5.05\%                  & -                  & \highlightRelevantValue{-6.81} & 10.87\%              & -                  & \highlightRelevantValue{-9.14} \\
Llama 3.1 70B        & 8.19\%                  & -                  & \highlightRelevantValue{-3.67} & 12.69\%              & -                  & \highlightRelevantValue{-7.32} \\
Mixtral 8x7B         & 5.41\%                  & -                  & \highlightRelevantValue{-6.45} & 13.16\%              & -                  & \highlightRelevantValue{-6.85} \\
Qwen2.5-7B           & 4.25\%                  & -                  & \highlightRelevantValue{-7.61} & 11.94\%              & -                  & \highlightRelevantValue{-8.07} \\
Qwen2.5-3B           & 3.91\%                  & -                  & \highlightRelevantValue{-7.95} & 10.87\%              & -                  & \highlightRelevantValue{-9.14} \\
Qwen2.5-1.5B         & 3.27\%                  & -                  & \highlightRelevantValue{-8.59} & 7.94\%               & -                  & \highlightRelevantValue{-12.07} \\
Mistral-7B-v0.3      & 4.25\%                  & -                  & \highlightRelevantValue{-7.61} & 11.27\%              & -                  & \highlightRelevantValue{-8.74} \\
Llama 3.2 3B         & 2.93\%                  & -                  & \highlightRelevantValue{-8.93} & 8.60\%               & -                  & \highlightRelevantValue{-11.41} \\
Llama 3.2 1B         & 3.71\%                  & -                  & \highlightRelevantValue{-8.15} & 3.09\%               & -                  & \highlightRelevantValue{-16.92} \\
Proprietary Models   &                         &                    &                  &                      &                    &                  \\
Claude 3.5 Haiku     & 9.18\%                  & -                  & \highlightRelevantValue{-2.68} & 14.77\%              & -                  & \highlightRelevantValue{-5.24} \\
Claude 3 Opus        & 6.78\%                  & -                  & \highlightRelevantValue{-5.08} & 15.08\%              & -                  & \highlightRelevantValue{-4.93} \\
Claude 3 Sonnet      & 8.42\%                  & -                  & \highlightRelevantValue{-3.44} & 17.40\%              & -                  & \highlightRelevantValue{-2.61} \\
Claude 3.5 Sonnet    & 9.72\%                  & -                  & \highlightRelevantValue{-2.14} & 15.91\%              & -                  & \highlightRelevantValue{-4.10} \\
Claude 3.5 Sonnet v2 & 11.69\%                 & -                  & \highlightRelevantValue{-0.17} & 18.54\%              & -                  & \highlightRelevantValue{-1.47} \\
Claude 3.7 Sonnet    & 9.34\%                  & -                  & \highlightRelevantValue{-2.52} & 12.81\%              & -                  & \highlightRelevantValue{-7.20} \\
Fine-tuned Models    &                         &                    &                  &                      &                    &                  \\
Qwen2.5-7B~          & \underline{16.67\%}                 & -                  & \highlightRelevantValue{4.81}  & 26.92\%              & -                  & \highlightRelevantValue{6.91}  \\
+ reasoning          & \textbf{17.26\%}                 & \highlightRelevantValue{3.54}  & \highlightRelevantValue{5.40}  & \underline{ 33.86\%}              & \highlightRelevantValue{25.78} & \highlightRelevantValue{13.85} \\
Qwen2.5-3B~          & 14.53\%                 & -                  & \highlightRelevantValue{2.67}  & 22.88\%              & -                  & \highlightRelevantValue{2.87}  \\
+ reasoning          & 11.88\%                 & \highlightRelevantValue{-18.24} & \highlightRelevantValue{0.02}  & 28.52\%              & \highlightRelevantValue{24.65} & \highlightRelevantValue{8.51}  \\
Qwen2.5-1.5B~        & 5.03\%                  & -                  & \highlightRelevantValue{-6.83} & 5.67\%               & -                  & \highlightRelevantValue{-14.34} \\
+ reasoning          & 16.06\%                 & \highlightRelevantValue{219.28} & \highlightRelevantValue{4.20}  & 27.69\%              & \highlightRelevantValue{388.36} & \highlightRelevantValue{7.68}  \\
Mistral-7B-v0.3~     & 14.17\%                 & -                  & \highlightRelevantValue{2.31}  & 17.99\%              & -                  & \highlightRelevantValue{-2.02} \\
+ reasoning          & 15.84\%                 & \highlightRelevantValue{11.79}  & \highlightRelevantValue{3.98}  & 30.12\%              & \highlightRelevantValue{67.43}  & \highlightRelevantValue{10.11} \\
Llama-3.2-3B~        & 9.31\%                  & -                  & \highlightRelevantValue{-2.55} & 4.73\%               & -                  & \highlightRelevantValue{-15.28} \\
+ reasoning          & 15.77\%                 & \highlightRelevantValue{69.39}  & \highlightRelevantValue{3.91}  & \textbf{33.99\%}              & \highlightRelevantValue{618.60} & \highlightRelevantValue{13.98} \\
Llama-3.2-1B~        & 11.13\%                 & -                  & \highlightRelevantValue{-0.73} & 10.44\%              & -                  & \highlightRelevantValue{-9.57} \\
+ reasoning          & 7.53\%                  & \highlightRelevantValue{-32.35} & \highlightRelevantValue{-4.33} & 15.08\%              & \highlightRelevantValue{44.44}  & \highlightRelevantValue{-4.93} \\
\bottomrule
\end{booktabs}

\caption{Performance comparison of models in different settings.}
\label{tab:results-full}
\end{table*}

\section{Prompts}

Reasoning Synthesize Prompt:
\begin{minted}[breaklines]{text}
You will be given a customer's shopping journey on one of the largest e-commerce platforms globally. you will be given the context (what the user is looking at), the action (what the user did), and your job is to predict the user's rationale for the action. The rationale should follow 
Here is an example:
{example}
For each action in the input, output a rationale.
If the action is "terminate", it means that you didn't find any desired product and you decided to leave the website by closing the browser window.
\end{minted}

Baseline model evaluation prompt:

\begin{minted}[breaklines]{markdown}
<IMPORTANT>
Your task is to predict the next action and provide rationale for the action based on the previous actions and context.
You need to pretend that you are a user, browsing one of the largest e-commerce platforms globally and searching for a product to purchase.
The history action (with details described below) and context will be provided to you.
You need to predict the next action and provide rationale for the action.
</IMPORTANT>


# Action Space

An action is represented in JSON format, and there are four primary types of actions:

#### 1. `type_and_submit`:
Type text into an input field and immediately submit the form. Equivalent to typing text into an input and pressing enter key.

{
    "type": "type_and_submit",
    "name": "input_name",
    "text": "search_text"
}


#### 2. `click`:
Click on a button or clickable element identified by `name`.


{
    "type": "click",
    "name": "clickable_name"
}


#### 3. `terminate`:
When you are unsatisfied with the current search result and you don't want to buy anything, use `terminate` to indicate that you want to close the browser window and terminate the task.

{
    "type": "terminate"
}

# Context
Your context will be an **simplified version** of the raw HTML of the one of the largest e-commerce platforms globally page you are looking at. Some interactable elements will be added a unique "name" attribute, which you can use to identify the element to interact with (click or type_and_submit).

# Rationale

The rationale is a first-person sentence of what you are thinking when you make the action. It should be a short sentence that explains why you are making the action.

# Output Format

You need to predict the next action and provide rationale for the action. Your output should follow a strict JSON form:

{
    "action": {
        // action goes here
        "type": "<type>",
        ...
    },
    "rationale": "<rationale>" // rationale goes here, a string
}

<IMPORTANT>
OUTPUT A SINGLE JSON OBJECT, NOTHING ELSE.
</IMPORTANT>
\end{minted}

\section{Example Context}
\begin{minted}{html}
<html>

<head>
    <title>Amazon.com : pressure washer</title>
</head>

<body>
    <form role="search"> <input name="search_input" value="pressure washer" type="text" aria-label="Search Amazon" />
        <input name="search_button" type="submit" value="Go" />
    </form>
    <div name="refinements">
        <div> <span class="refinement-title"> Amazon Prime </span>
            <li name="refinements.amazon_prime.prime_eligible" role="checkbox"> Prime Eligible <input type="checkbox">
            </li>
        </div>
        <div> <span class="refinement-title"> Prime Delivery </span>
            <li name="refinements.prime_delivery.tomorrow_by_8am" role="checkbox"> Tomorrow by 8AM <input
                    type="checkbox"> </li>
        </div>
        <div> <span class="refinement-title"> Delivery Day </span>
            <li name="refinements.delivery_day.get_it_today" role="checkbox"> Get It Today <input type="checkbox"> </li>
            <li name="refinements.delivery_day.get_it_by_tomorrow" role="checkbox"> Get It by Tomorrow <input
                    type="checkbox"> </li>
        </div>
        <div> <span class="refinement-title"> Customer Reviews </span>
            <li name="refinements.customer_reviews.4_stars__up" role="checkbox"> 4 Stars &amp; Up <input
                    type="checkbox"> </li>
        </div>
        <div> <span class="refinement-title"> All Top Brands </span>
            <li name="refinements.all_top_brands.top_brands" role="checkbox"> Top Brands <input type="checkbox"> </li>
        </div>
        <div> <span class="refinement-title"> Deals &amp; Discounts </span>
            <li name="refinements.deals__discounts.all_discounts" role="checkbox"> All Discounts <input type="checkbox">
            </li>
            <li name="refinements.deals__discounts.todays_deals" role="checkbox"> Today&#39;s Deals <input
                    type="checkbox"> </li>
        </div>
        <div> <span class="refinement-title"> Pressure Washer Pressure </span>
            <li name="refinements.pressure_washer_pressure.under_1700_psi" role="checkbox"> Under 1700 PSI <input
                    type="checkbox"> </li>
            <li name="refinements.pressure_washer_pressure.1700_to_1999_psi" role="checkbox"> 1700 to 1999 PSI <input
                    type="checkbox"> </li>
            <li name="refinements.pressure_washer_pressure.2000_to_2599_psi" role="checkbox"> 2000 to 2599 PSI <input
                    type="checkbox"> </li>
            <li name="refinements.pressure_washer_pressure.2600_to_2799_psi" role="checkbox"> 2600 to 2799 PSI <input
                    type="checkbox"> </li>
            <li name="refinements.pressure_washer_pressure.2800_to_3099_psi" role="checkbox"> 2800 to 3099 PSI <input
                    type="checkbox"> </li>
            <li name="refinements.pressure_washer_pressure.3100_to_3999_psi" role="checkbox"> 3100 to 3999 PSI <input
                    type="checkbox"> </li>
            <li name="refinements.pressure_washer_pressure.4000_psi__above" role="checkbox"> 4000 PSI &amp; Above <input
                    type="checkbox"> </li>
        </div>
        <div> <span class="refinement-title"> Condition </span>
            <li name="refinements.condition.renewed" role="checkbox"> Renewed <input type="checkbox"> </li>
            <li name="refinements.condition.new" role="checkbox"> New <input type="checkbox"> </li>
            <li name="refinements.condition.used" role="checkbox"> Used <input type="checkbox"> </li>
        </div>
        ...
        <div> <span class="refinement-title"> Color </span>
            <li name="refinements.color.black" role="checkbox"> Black <input type="checkbox"> </li>
            <li name="refinements.color.multi" role="checkbox"> Multi <input type="checkbox"> </li>
            <li name="refinements.color.blue" role="checkbox"> Blue <input type="checkbox"> </li>
            <li name="refinements.color.grey" role="checkbox"> Grey <input type="checkbox"> </li>
            <li name="refinements.color.white" role="checkbox"> White <input type="checkbox"> </li>
            <li name="refinements.color.brown" role="checkbox"> Brown <input type="checkbox"> </li>
            <li name="refinements.color.beige" role="checkbox"> Beige <input type="checkbox"> </li>
            <li name="refinements.color.red" role="checkbox"> Red <input type="checkbox"> </li>
            <li name="refinements.color.pink" role="checkbox"> Pink <input type="checkbox"> </li>
            <li name="refinements.color.orange" role="checkbox"> Orange <input type="checkbox"> </li>
            <li name="refinements.color.yellow" role="checkbox"> Yellow <input type="checkbox"> </li>
            <li name="refinements.color.ivory" role="checkbox"> Ivory <input type="checkbox"> </li>
            <li name="refinements.color.green" role="checkbox"> Green <input type="checkbox"> </li>
            <li name="refinements.color.purple" role="checkbox"> Purple <input type="checkbox"> </li>
            <li name="refinements.color.gold" role="checkbox"> Gold <input type="checkbox"> </li>
            <li name="refinements.color.silver" role="checkbox"> Silver <input type="checkbox"> </li>
            <li name="refinements.color.clear" role="checkbox"> Clear <input type="checkbox"> </li>
        </div>
        <div> <span class="refinement-title"> More-sustainable Products </span>
            <li name="refinements.moresustainable_products.climate_pledge_friendly" role="checkbox"> Climate Pledge
                Friendly <input type="checkbox"> </li>
        </div>
    </div>
    <div>
        <div class="search-result"> <a
                name="search_results.2025_upgraded_electric_pressure_washer_with_adjustable_touch_screen_5000_psi_33_gpm_8_power_settings_for_car_patio__floor_cleaning_detergent_tank_with_4_nozzles_23ft_hose_35ft_cord.view_product"
                class="product-name"> 2025 Upgraded Electric Pressure Washer with Adjustable Touch Screen, 5000 PSI
                3.3
                GPM, 8 Power Settings, for Car, Patio &amp; Floor Cleaning, Detergent Tank, with 4 Nozzles, 23ft
                Hose,
                35ft Cord </a>
            <div class="product-review"> <span class="product-rating">5.0 out of 5 stars</span><span
                    class="product-rating-count"> 20.0 reviews </span> </div>
            <div class="product-price"><span>$199.99</span></div>
            <div class="product-delivery">FREE delivery Mon Mar 17</div>
        </div>
        <div class="search-result"> <a
                name="search_results.2025upgraded_electric_pressure_washer_4500_psi_32_gpm_power_washer_with_4_quick_connect_nozzles_inlet_hose__filter_foam_cannon_for_carsfencesdrivewayshome_cleaning.view_product"
                class="product-name"> 2025Upgraded Electric Pressure Washer, 4500 PSI 3.2 GPM Power Washer with 4
                Quick
                Connect Nozzles, Inlet Hose &amp; Filter&amp; Foam Cannon for Cars/Fences/Driveways/Home Cleaning
            </a>
            <div class="product-review"> <span class="product-rating">5.0 out of 5 stars</span><span
                    class="product-rating-count"> 19.0 reviews </span> </div>
            <div class="product-price"><span>$159.99</span></div>
            <div class="product-delivery">FREE delivery Mon Mar 17</div>
        </div>
        <div class="search-result"> <a
                name="search_results.kärcher_pressure_washer_k1800ps_max_2250_psi_3_spray_nozzles_detergent_tank_for_cars_driveways_siding_patios_146_max_gpm.view_product"
                class="product-name"> Kärcher Pressure Washer K1800PS, Max 2250 PSI, 3 Spray Nozzles, Detergent
                Tank,
                For Cars, Driveways, Siding, Patios, 1.46 max. GPM </a>
            <div class="product-review"> <span class="product-rating">4.3 out of 5 stars</span><span
                    class="product-rating-count"> 8520.0 reviews </span> </div>
            <div class="product-price"><span>$167.99</span></div>
        </div>
        <div class="search-result"> <a
                name="search_results.kärcher_pressure_washer_k1700_max_2125_psi_3_spray_nozzles_detergent_tank_for_cars_driveways_siding_patios_146_max_gpm.view_product"
                class="product-name"> Kärcher Pressure Washer K1700, Max 2125 PSI, 3 Spray Nozzles, Detergent Tank,
                For Cars, Driveways, Siding, Patios, 1.46 max. GPM </a>
            <div class="product-review"> <span class="product-rating">4.3 out of 5 stars</span><span
                    class="product-rating-count"> 8520.0 reviews </span> </div>
            <div class="product-price"><span>$145.99</span></div>
        </div>
        <div class="search-result"> <a
                name="search_results.kärcher_pressure_washer_k2300ps_max_2875_psi_4_spray_nozzles_detergent_tank_hose_reel_for_cars_driveways_siding_patios_207_max_gpm.view_product"
                class="product-name"> Kärcher Pressure Washer K2300PS, Max 2875 PSI, 4 Spray Nozzles, Detergent
                Tank, Hose Reel, For Cars, Driveways, Siding, Patios, 2.07 max. GPM </a>
            <div class="product-review"> <span class="product-rating">4.3 out of 5 stars</span><span
                    class="product-rating-count"> 8520.0 reviews </span> </div>
            <div class="product-price"><span>$223.98</span></div>
        </div>
        <span name="pagination">
            <span name="pagination.1" aria-label="Current page, page 1">1</span>
            <a name="pagination.2" aria-label="Go to page 2">2</a>
            <a name="pagination.3" aria-label="Go to page 3">3</a>
            <a aria-label="...">...</a>
            <a name="pagination.7" aria-label="Go to page 7">7</a>
        </span>
    </div>
</body>

</html>
\end{minted}

\end{document}